\documentclass{article}

\usepackage{PRIMEarxiv}

\usepackage[utf8]{inputenc} 
\usepackage[T1]{fontenc}    
\usepackage{hyperref}       
\usepackage{url}            
\usepackage{booktabs}       
\usepackage{amsfonts,amsmath}       
\usepackage{nicefrac}       
\usepackage{microtype}      
\usepackage{lipsum}
\usepackage{fancyhdr}       
\usepackage{graphicx}       
\graphicspath{{media/}}     
\usepackage[linesnumbered,ruled]{algorithm2e}
\usepackage{url,bm,bbm,multicol,multirow}
\usepackage{algorithmic,float,amssymb}
\usepackage[normalem]{ulem}
\usepackage[export]{adjustbox}
\usepackage{epstopdf} 
\usepackage{xcolor}
\usepackage{times}

\newtheorem{lemma}{Lemma}

\newtheorem{definition}{Definition}

\newcommand{\E}{\mathbb{E}}
\DeclareMathOperator*{\argmin}{arg\,min}

\let\classAND\AND
\let\AND\relax
\usepackage{algorithmic}

\let\AND\classAND
\AtBeginEnvironment{algorithmic}{\let\AND\algoAND}
\title{Leveraging Black-box Models to Assess\\ 
Feature Importance in {Unconditional} Distribution
}

\author{Jing Zhou\\
      School of Engineering, Mathematics, and Physics\\
      University of East Anglia\\
       \texttt{J.Zhou6@uea.ac.uk}\\
      \AND
       Chunlin Li \\
       Department of Statistics\\
      Iowa State University \\
  \texttt{chunlin@iastate.edu}
}

\begin{document}
\maketitle

\begin{abstract}
Understanding how changes in explanatory features affect the unconditional distribution of the outcome is important in many applications.
However, existing black-box predictive models are not readily suited for analyzing such questions. 
In this work, we develop an approximation method to compute the feature importance curves relevant to the unconditional distribution of outcomes, while leveraging the power of pre-trained black-box predictive models. The feature importance curves measure the changes across quantiles of outcome distribution given an external impact of change in the explanatory features. Through extensive numerical experiments and real data examples, we demonstrate that our approximation method produces sparse and faithful results, and is computationally efficient. 
\end{abstract}


\section{Introduction}

In recent years, black-box models, particularly those based on complex machine learning algorithms such as deep learning and ensemble methods, have demonstrated remarkable success in predictive tasks. These models, including popular pre-trained architectures like transformers and large models, have reshaped fields such as natural language processing, computer vision, and structured data analysis. Their strength lies in their ability to model \emph{conditional} distributions: given a set of features, they predict a target outcome with high accuracy. This capability has led to widespread adoption across industries for tasks such as classification, regression, and forecasting. The predictive power of these black-box models often surpasses traditional statistical approaches, making them a cornerstone of modern big data applications.

However, in many applications, the focus shifts away from merely predicting outcomes based on features, toward understanding how changes in these features affect the \emph{marginal} (or \emph{unconditional}) distribution of the outcome. For instance, in policymaking, fairness and equity, or program evaluation, we may be more interested in identifying how specific features influence different segments of the outcome population rather than predicting individual outcomes. Take the study of income distribution as an example: we may not only care about predicting an individual's income based on their characteristics but also about understanding which features have the most significant impact on the lower 50\% of the \emph{overall} income distribution. This shift in focus reveals that not every feature is equally important across the entire outcome distribution, and certain features may only play a critical role in specific quantiles or regions. So far, most existing predictive models, particularly pre-trained ones, are designed to optimize conditional distributions and are not readily suited for analyzing such questions. This raises a crucial question: 
\begin{center}
    {\it Can we leverage the predictive capacity of these black-box models to assess feature importance relevant to the unconditional distribution of outcomes?}
\end{center}

In this paper, we provide a definitive answer to this question and develop an estimation procedure for estimating the feature importance curve across different quantiles of the unconditional distribution of the outcome, 
without rebuilding/fitting black-box model.

\subsection{Our Contributions}

Specifically, this paper contributes to the following four aspects.

We take a closer look at the problem of measuring feature importance relevant to the outcome's marginal distribution, under the framework of the \emph{unconditional quantile regression} (UQR) \cite{firpo2009unconditional}. Then, we present an effective and efficient strategy to approximate the key quantities in the feature importance score, leveraging the output of a predictive black-box model. Unlike the existing methods in the UQR literature, which requires fitting new models, our approach does not require retraining any complex predictive model when applied to identically distributed data. If the data distribution is slightly shifted, it also allows fine-tuning the existing pretrained model to maximize knowledge transfer; this will become clear in Section~\ref{ssec:main-method}.

It will become evident that our novel importance measure in Eq.~(\ref{equation:feature-importance}) is \emph{post-hoc} \cite{athalye2018explaining,doshi-velez2017model}. It can be combined with any pre-trained black-box algorithms estimating the association between features and outcome, including neural networks. The importance measure is computationally efficient. Technically, we use the technique called \emph{density extrapolation} to address the challenge of nonparametric estimation for the tail part of the error density function. 

Additionally, we propose a fast pruning procedure based on the asymptotic properties of the influence function; the pruning does not require retraining the model. The pruning procedure is automatically activated only if the pretrained model is sufficiently accurate with respect to $h(\cdot)$. The pruning sparsifies the feature importance measure and unifies feature importance with feature selection to enhance the interpretation of the pretrained models.

Through comprehensive empirical studies, we confirm that the proposed method delivers faithful and sparse feature importance in various complex data scenarios, including nonparametric regression and high-dimensional regression.

\subsection{Related Work}

This work has connections to the following areas.

\paragraph{Quantile Regression.} 
As a key tool in statistical analysis, quantile regression can be broadly categorized into conditional quantile regression and unconditional quantile regression. Conditional quantile regression \cite{koenker2005quantile} focuses on estimating quantiles of a response variable conditional on specific values of explanatory variables, while unconditional quantile regression \cite{firpo2009unconditional} aims to estimate quantiles of the marginal distribution of the response variable. Our work utilizes the latter to assess feature importance across different unconditional (marginal) quantiles.
Conventional approaches \cite{firpo2009unconditional} to unconditional quantile regression often rely on the estimation of conditional probabilities $P(Y>q_{\tau} \mid X=x)$, typically achieved by building a dedicated model trained on the dataset of interest. However, this process poses challenges when leveraging the capabilities of pretrained black-box models or incorporating external information. The dependence on explicitly training a model not only limits the adaptability to diverse pretrained architectures but also constrains the ability to integrate auxiliary data seamlessly. Our method addresses these limitations, enabling a more flexible and robust analysis of feature importance in the context of unconditional quantile estimation.

\paragraph{Sparse Learning and Model agnostic Feature Importance for explaining Black-Box models.} 

To provide interpretation for black-box models, two aspects can be addressed: (1) feature importance: measuring the importance of the features to prediction; (2) sparse learning: automatically deciding a subset of features entering the black-box model and separating the collection of features into ``relevant" and ``irrelevant" to prediction. Some widely accepted model agnostic approaches measuring features importance in machine learning include the Shapley value \cite{shapley1953stochastic,roth1988introduction} and permutation importance \cite{molnar2019interpretable,doshi-velez2017model,athalye2018explaining}. The first measures the importance of each features by considering the contribution of it in all possible submodels, whereas the latter compares the performance of the model before and after shuffling the values of features. Such feature importance measures are often computationally intensive and requires retraining the models, which is challenging when the number of features is large. 
To make use of the power of machine learning black-box algorithms, sparse learning can be considered. Different strategies such as best subset selection, forward selection, backward elimination, regularization can be considered, see \cite{guyon2003introduction} for a review of variable selection in machine learning.

Our proposed method allows for a fast computation of model agnostic feature importance measure. Based on importance measures, we further automatically pruning the importance measure in order to obtain a subset of ``relevant" features contributing to predicting the outcome at different quantile levels.

\paragraph{Transfer Learning and Causal Invariance Principle.}  
Our approach relies on an important assumption: the conditional distribution of the outcome given the features is invariant to perturbations in the marginal distributions of the features. This assumption is common in transfer learning and causality, as it guarantees that the structures remain the same across new datasets or environments.

\section{Problem Formulation}

\begin{figure}
    \centering
    \includegraphics[width=0.5\linewidth]{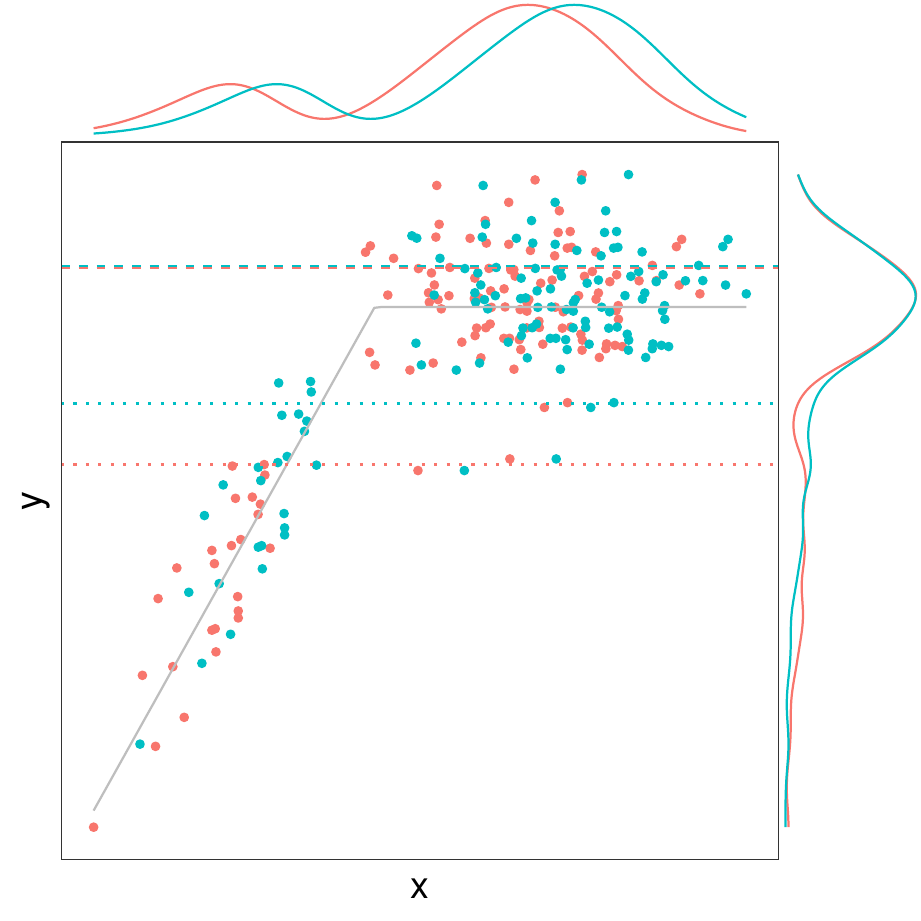}
    \caption{Illustration of $\beta(\tau)$. Red points are the original distribution of $(X, Y)$. Blue points are the counterfactual distribution after a shift intervention on $X$. Grey line is $\E(Y\mid X)$. $\beta(0.2) > 0$ but $\beta(0.8)\approx 0$.} 
    \label{fig:illustration}
\end{figure}

We evaluate feature importance to the outcome's marginal distribution under the framework of UQR \cite{firpo2009unconditional}. 
Let $Y\in\mathbb R$ be a (continuous) outcome variable and $X\in\mathbb R^p$ be explanatory features, such that $(X, Y)\sim F_{X, Y}$. 
Mimicking the interpretation of classical regression coefficients, we consider an intervention causing a small location shift in the distribution of $X$ and measure the change in the unconditional distribution of $Y$; see Figure \ref{fig:illustration} for an illustration.

To proceed, let $F_X$, $F_Y$, $\widetilde F_X$, and $\widetilde F_Y$ be the observed marginal distributions of $X,Y$ and the counterfactual marginal distributions of $X,Y$ after the intervention, respectively. Assume that the conditional distribution $F_{Y\mid X}$ is invariant to such an intervention on $X$. Then 
\begin{equation}\label{equation:distribution}
F_{Y}(y) = \int F_{Y\mid X}(y\mid X=x) d F_X(x),
\qquad 
\widetilde F_{Y}(y) = \int F_{Y\mid X}(y\mid X=x) d \widetilde F_X(x).
\end{equation}
We use \emph{marginal} quantiles $\{q_{\tau}\}_{\tau\in (0,1)}$ to characterize the change in the unconditional distribution of $Y$. 
For $\tau\in(0,1)$, consider the $\tau$-th quantile $q_{\tau}=Q_{\tau}(F_Y)\equiv \inf\{ y\in\mathbb R : F_Y(y)\geq \tau \}$, where $Q_{\tau}$ is a functional mapping the distribution $F_Y$ to $q_{\tau}$. 
Now, suppose we impose a small perturbation on the features, causing a local shift from $X$ to $X+t$. Let $\widetilde F_X = F^{(t)}_X$ be the distribution of $X+t$, and let $F_Y^{(t)}$ be the corresponding distribution of $Y$ using $F^{(t)}_X$ and $F_{Y\mid X}$ as in Eq.~(\ref{equation:distribution}).
Then, it follows from the von Mises linear approximation (a distributional analog of the Taylor series) that 
\begin{equation}\label{equation:feature-importance}
    \begin{split}
        \beta(\tau) \equiv \frac{d}{dt}Q_{\tau}(F^{(t)}_Y ) \Big|_{t=0}
    &= \int  \frac{\partial }{\partial x}\mathbb E\Big[\text{IF}_{Q_{\tau},F_Y}(Y) \mid X=x\Big] d F_X(x)\\
    &= \frac{1}{f_Y(q_{\tau})}\int \frac{\partial }{\partial x} \mathbb P\Big[Y > q_\tau \mid X = x\Big] d F_X(x),
    \end{split} 
\end{equation}
where $\mbox{IF}_{Q_{\tau}, F_Y}(y) = (\tau - \mathbbm 1[y\leq q_{\tau}])/ f_Y(q_{\tau})$ is the influence function, and $f_Y$ is the density of $Y$; see Appendix \ref{sec:appendix} for a derivation.

Thus, $\beta(\tau)$ serves as a $p$-dimensional feature importance vector, measuring the contribution of each feature to the change in the unconditional $\tau$-th quantile of $Y$ under an infinitesimal location shift. 
Our goal is to estimate the $p$-dimensional function $\beta(\tau)$ with $\tau\in(0,1)$.

In most, if not all, UQR literature, modeling has focused on $\mathbb{P}(Y > q_{\tau} \mid X)$. However, this approach requires dichotomizing the outcome and fitting a classification model to estimate the conditional probability. When a good predictive model, $\hat{h}(x)$, is already available, such an approach becomes undesirable. Dichotomizing the outcome is, in some sense, equivalent to masking $h(\cdot)$ with an additional unknown function, which moves in the opposite direction of our goal—interpreting a pre-trained black-box model.

\section{Method}

In this section, we present a general framework for computing the approximation of $\beta(\cdot)$. To leverage the information of a predictive model, the key is to model the conditional distribution $Y\mid X$ in a fashion that accommodates the flexible black-box algorithm. 

\subsection{Method for Black Box with Univariate Output}\label{ssec:main-method}

We commence with a simple yet common scenario: 
suppose we can access a black-box predictive model $\widehat h$ with a univariate output, predicting the (continuous) outcome $Y\in\mathbb R$ based on the input $X\in\mathbb R^p$. 
Without further information, the conditional distribution $Y\mid X$ is not specified. Thus, we assume the data are randomly sampled from 
\begin{equation}\label{equation:nonlinear-model}
    Y = h(X, \varepsilon),
\end{equation}
where $h$ is the unknown truth function that $\widehat h$ aims to learn, and $\varepsilon$ is some unobserved error independent of $X$. The functional form allows for interaction between the features and the error term. In the case of using popular neural networks to learn $h$, we allow for fine-tuning the output layer $\widehat g_2$ while fixing $\widehat g_1$ in ${\widehat h} = \widehat g_1 \circ \widehat g_2 $ in order to better adapt to new datasets with potential marginal distribution shifts. In many applications, Eq.~(\ref{equation:nonlinear-model}) is a reasonable approximation, as there is typically a low-dimensional relationship (e.g., linear) between the observed outcome $Y$ and predicted outcome $\widehat Y=\widehat h(X)$ \cite{wang2020methods}. 

Often, there is a gap between $Y$ and $\widehat Y  = \widehat h(X, \varepsilon)$. We denote 
\begin{equation}\label{equation: residual define}
    Y  = \widehat h(X, \varepsilon) + R,
\end{equation}
where the residual $R$ measures the distance between the observed and predicted outcomes. In addition, independence holds between $R $ and $X$ if $\widehat h$ well-approximates the association between $X$ and $Y$. 

Now, combining Eq.~(\ref{equation:feature-importance}) and Eq.~(\ref{equation: residual define}) yields 
\begin{equation}\label{equation:feature-importance-univariate-output}
    \begin{split}
        \beta(\tau) 
        &= \mathbb E_{X\sim F_X} \left[ \frac{1}{f_Y(q_{\tau})} \frac{\partial}{\partial x} \Big(1 - \mathbb P\Big[ R\leq q_{\tau} - \widehat h(X, \varepsilon)\Big]\Big)\right] \\
        &= \mathbb E_{X\sim F_X} \left[ \frac{f_{R}(q_{\tau} - \widehat h(X, \varepsilon))}{f_Y(q_{\tau})} \frac{\partial \widehat h(X, \varepsilon)}{\partial x}\right],
    \end{split}
\end{equation}
where the partial derivative w.r.t $x$ represents the contribution of the features to the conditional mean of $Y$ and the scaling factor $\frac{f_{R}(q_{\tau} - \widehat h(X, \varepsilon))}{f_Y(q_{\tau})}$ depends on quantile level $\tau$.

This measure is suitable for post-hoc analysis of a pre-trained model with two major advantages. Computationally, this new measure requires simple calculation without considering $2^p$ combinations or retraining model as opposed to SHAP and permutation importance. This is especially attractive when $p$ is large. In addition, this measure allows flexibility of user's choice of algorithms estimating $h(x, \varepsilon)$; this is different from measures such as random forest \cite{breiman2001random}.

\paragraph{Estimating $\beta(\tau)$ and Challenges.} Eq.~(\ref{equation:feature-importance-univariate-output}) suggests a natural plug-in estimator for $\beta(\tau)$, 
\begin{equation}\label{equation:estimator-univariate-output}
\widehat\beta(\tau) = \frac{1}{n}\sum_{i = 1}^n
   \frac{ \widehat f_R(\widehat q_{\tau} - \widehat h(X_i, \varepsilon_i))}{\widehat f_Y(\widehat q_\tau)} \frac{\partial \widehat h(X_i, \varepsilon_i)}{\partial x},
\end{equation}
where $\widehat q_{\tau}$ is the estimator for the $\tau$-th unconditional quantile of $Y$, and $\widehat f_{Y}$ and $\widehat f_\varepsilon$ are density estimators of $Y$ and $\varepsilon$, respectively.
As such, as long as $\tau\not\to 0$ or $\tau\not\to 1$, the quantities $\widehat q_{\tau}$ and $\widehat f_Y(\widehat q_{\tau})$ can be well estimated by conventional estimators
\begin{equation}\label{equation:empirical-quantile}
    \widehat q_\tau = \argmin_{q \in \mathbb R} \sum_{i =1}^n (\tau - \mathbbm 1 [Y_i -q \le 0]) (Y_i - q),
\end{equation}
\begin{equation}\label{equation:kernel-density-at-q_tau}
\widehat f_Y(\widehat q_\tau) = \frac{1}{n b_Y} \sum_{i =1}^n K_Y \left(\frac{Y_i - \widehat q_\tau}{b_Y}\right),
\end{equation}
with $K_Y$ being some kernel function and $b_Y$ being the bandwidth. 

In contrast, estimating $\widehat f_{\varepsilon}(\widehat q_{\tau} - \widehat h(X_i, \varepsilon_i))$ is much more challenging. To see this, ignoring the estimation error so that $\widehat q_{\tau}\approx q_{\tau}$ and $\widehat h\approx h$, then the error density is evaluated at points $q_\tau - h(X_i, \varepsilon_i)$, which may be values beyond the empirical range of $R_i$,
even if $\tau\not\to 0$ and $\tau\not\to 1$. Figure \ref{fig:illustration} provides an illustration. 
This highlights the pressing need to devise a new density estimator. 
Additionally, $\widehat\beta(\tau)$ in Eq.~(\ref{equation:estimator-univariate-output}) does not yield a sparse feature importance vector to exclude non-informative features automatically. A sparse measure is highly desirable in practical data analysis, particularly when the number of explanatory features is large.
We will address the two issues for $\widehat\beta(\tau)$ in Section~\ref{ssec: error density extrapolate} and \ref{ssec: backward elimination}, respectively.

\subsubsection{Error Density Estimation via Extrapolation}\label{ssec: error density extrapolate}

\begin{figure}[!hbt]
    \centering
    \includegraphics[width=0.5\linewidth]{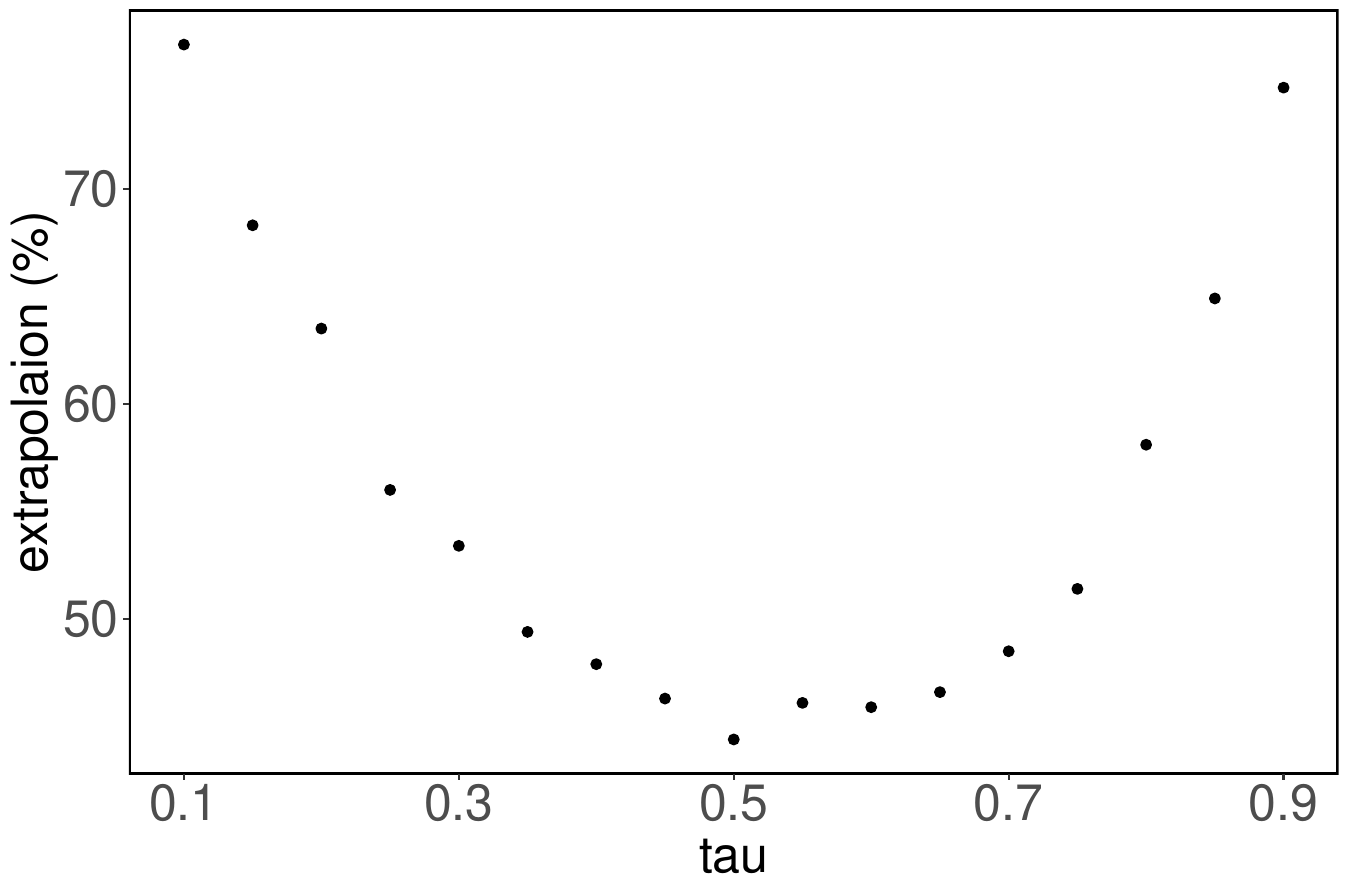}
    \caption{The percentage of out-of-range observations against quantile level $\tau$. The data are generated from $Y = 1 -2 X_1 + 5X_2 + \varepsilon$, where $\varepsilon \sim N(0,1)$, $X = (X_1, \ldots, X_4)^\top \sim N(\bm 0_4, \Sigma)$, and $\Sigma_{ij} = 0.5^{i - j}$.}
    \label{fig:extrapolation}
\end{figure}

Let $\widehat R_i = Y_i - \widehat h(X_i, \varepsilon_i)$, 
the challenge of estimating $\widehat f_{R}(\widehat q_{\tau} - \widehat h(X_i, \varepsilon_i))$ arises from that $\widehat q_{\tau} - \widehat h(X_i, \varepsilon_i)$ may not be in the interval $\big[\min_{i} \widehat R_i, \max_i \widehat R_i\big]$, as illustrated in Figure~\ref{fig:extrapolation}. In this simple linear regression example, the minimum extrapolation percentage, around 45\%, occurs at $\tau = 0.5$. When $\tau$ deviates from median, ``out-of-range"  occurrences become more frequent, and this issue becomes more severe as $\tau$ approaches the tail quantiles. For such data points, there is limited information from the sample, causing unreliable estimation. 
To address this issue, we use a popular technique in extreme value theory, called density extrapolation, by assuming that distribution $F_R$ has regularly varying tails 
  \begin{equation}\label{eq: regularly varying tails}
   1- F_R(r) = L(r) r^{-1/\gamma}, 
  \end{equation}
 where $L(r) $ is a slowly varying function at $\infty$, that is, $\lim_{r \to \infty}\frac{L(ar)}{L(r)} = 1$ for all $a >0$. The tail index $\gamma >0$ covers the heavy-tailed distributions and $\gamma = 0$ covers the light-tailed distributions. Typically, the extrapolation considers estimating at quantile level $1-\tau_n \to 1$ with effective sample size $n\tau_n \to \infty $, then use the tail information in Eq.~(\ref{eq: regularly varying tails}) to extrapolate to regions where there is limited or no information. 

 To avoid estimating the nuisance slowly varying function, we consider the following ratio 
 \begin{equation}\label{eq: error ratios}
  \lim_{r\to \infty}\frac{f_\varepsilon(ar)}{f_\varepsilon(r)} = \lim_{r\to \infty} \frac{1 - F_\varepsilon(ar)}{1 - F_\varepsilon(r)} = \lim_{r\to \infty} \frac{L(ar) (ar)^{-1/\gamma}}{L(r) r^{-1/\gamma} } = \lim_{r\to \infty}\frac{L(ar)}{L(r)} a^{-1/\gamma} = a^{-1/\gamma},
 \end{equation}
where the first equality holds by L'H\^opital's rule, the second equality holds by \eqref{eq: regularly varying tails}, the last equality holds by using $L(\cdot)$ is a slowly varying function. By taking $a = r'/r$ where $r$ is a prechosen error tail quantile $r = q_{\varepsilon, 1-\tau_n}$ satisfying the order $n\tau_n \to \infty$ and $r'$ being the out-of-range values, the density extrapolation based on Eq.~(\ref{eq: error ratios}) is as follows 
\begin{equation}
    f_\varepsilon(r') = (\frac{r'}{r})^{-1/{\gamma}} f_\varepsilon(r). 
\end{equation}
Based on this, we propose to estimate $\widehat f_{R}$ by
\begin{equation}\label{eq: kernel density extrapolate}
    \begin{split}
        \widehat f_R(\widehat q_\tau - \widehat h(X_i, \varepsilon_i)) 
        &= \frac{1}{n b_{1,R}}\sum_{i=1}^n K_R\left(\frac{\widehat R_{i'} - (\widehat q_\tau - \widehat h(X_i, \varepsilon_i))}{b_{1,R}}\right) \mathbbm 1\Big[\widehat R_{(n)} \geq \widehat q_\tau - \widehat h(X_i, \varepsilon_i) \Big]\\
        & 
        + \left(\frac{\widehat q_\tau - \widehat h(X_i, \varepsilon_i)}{\hat q_{R, 1-\tau_n}}\right)^{-1/\hat\gamma}
 \frac{1}{n b_{2,R}}\sum_{i=1}^n K_R\left(\frac{\widehat R_{i} - \widehat q_{R, 1-\tau_n}}{b_{2,R}}\right) \mathbbm 1\Big[\widehat R_{(n)} < \hat q_\tau - \widehat h(X_i, \varepsilon_i)\Big],
    \end{split}
\end{equation}
where 
\begin{equation}\label{eq: quantile estimator}
    \widehat q_{R, 1-\tau_n} = \argmin_{q \in \mathbb R} \sum_{i=1}^n \left(1- \tau_n -\mathbbm 1\Big[\widehat R_{i} - q \le 0\Big]\right) (\widehat R_{i} - q),
\end{equation}
and $\hat\gamma$ is the classic Hill estimator 
\begin{equation}\label{eq: Hill estimator}
    \widehat \gamma = \frac{1}{n\tau_n} \sum_{i=1}^n \Big(\log(\widehat R_{i'}) - \log(\widehat q_{R,1-\tau_n}) \Big) \mathbbm 1\Big[\widehat R_{i} > \widehat q_{R,1-\tau_n} \Big].
\end{equation}

\subsubsection{Sparsified Feature Importance via  Stepwise Backward Pruning}\label{ssec: backward elimination}



In Eq.~(\ref{equation:estimator-univariate-output}), $\widehat{\bm\beta}(\cdot)$ often contains near-zero entries, indicating that the corresponding features are almost irrelevant, if at all, to explaining $Y$. Such near-zero estimates can weaken the interpretability of $\widehat{\bm\beta}(\cdot)$, as it becomes difficult to determine whether they result from estimation errors or genuinely explain the variation in $Y$. Therefore, it is desirable to sparsify the feature importance scores, enhancing interpretability by pruning features that are irrelevant to $Y$.


To this end, we aim to identify a subset $\mathcal{S} \subseteq {1, \ldots, p}$ such that $Y$ is conditionally independent of $\bm{X}_{\mathcal{S}^c}$ given $\bm{X}_{\mathcal{S}}$, where $\mathcal{S}^c$ denotes the complement of $\mathcal{S}$. In this case, the features $\bm{X}_{\mathcal{S}^c}$ are irrelevant to explaining $Y$ and will be dropped by setting $\widehat{\bm\beta}_{\mathcal S^c}(\cdot) = \bm 0_{p-|\mathcal S|}$.

Pruning is carried out using an adaptive hard-thresholding technique for $\widehat{\bm{\beta}}(\tau)$ at a set of gridded quantile levels $\tau \in \{\tau_1, \ldots, \tau_K\}$. When the grid is sufficiently fine and $\bm{\beta}(\cdot)$ is smooth, $\widehat{\beta}_j(\tau) \approx 0$ for $\tau \in {\tau_1, \ldots, \tau_K}$ implies that $\beta_j(\cdot) \approx 0$. A key challenge is to adaptively select the thresholds across the different quantile levels.

The hard-thresholding procedure is not based directly on $\widehat{\beta}(\tau)$ since the baseline $\beta(\tau)$ is unknown. Instead, we focus on $\hat{q}_\tau(F_Y)$, for which the asymptotic normality result in Eq.~(\ref{eq: asymptotic normality}) holds.
Our goal is to identify a subset $\mathcal{S}^c$ to remove such that its exclusion does not result in significant deviations in $q_\tau(F_Y)$ at any quantile level $\tau = \tau_1, \ldots, \tau_K$. This adaptive hard-thresholding procedure selects the threshold at each quantile level $\tau = \tau_k$ based on the importance of the last feature whose removal causes minimal deviation, as measured by Eq.~(\ref{eq: asymptotic normality}).

We are interested in dropping a subset $\mathcal{S}^c$ such that its removal does not cause large deviations in $q_\tau(F_Y)$ across all quantile levels $\tau = \tau_1, \ldots, \tau_K$. This selection involves adaptive hard-thresholding for $\widehat{\bm\beta}(\tau)$ at multiple quantile levels. The threshold at each quantile level $\tau = \tau_k$ is determined by the estimated importance of the last feature whose removal causes negligible deviation. The deviation is measured as in Eq.~(\ref{eq: asymptotic normality}).



 
 To allow approximate $q_{\tau}(F_Y)$ using the features, we introduce
 \begin{eqnarray}\label{eq: recentered influence function}
   q_\tau(\delta_{y}) = q_\tau + \mbox{IF}(y; q_\tau) + o(1) = q_\tau + \frac{\tau - I\{y \leq q_\tau\}}{f_Y(q_\tau)} + o(1).
\end{eqnarray}

We denote the cardinality of the set $\mathcal S$ by $s$, a $(p-s)$-vector of all zeros by $\bm 0_{p-s}$. Further, we denote the subvector with variables $X_j \in S$ being $X_S$ and the subvector with $X_j \in \mathcal S^c$ being $X_{\mathcal{S}^c}$.
    By \eqref{eq: recentered influence function} and using the conditional independence $F_{Y\mid X} = F_{Y\mid X_{\mathcal S}} = F_{Y\mid (X_{\mathcal S}, \tilde X_{\mathcal S^c})}$, where $\tilde X_{\mathcal S^c, j}, j = 1, \ldots, s^c$ have a point mass at 0 with density function $F_{\tilde X_{\mathcal S^c, j}}(x) = \delta_0 = I\{\tilde X_{\mathcal S^c, j} = 0\}$, we obtain the following two ways of decomposing $q_\tau(F_Y)$
\begin{eqnarray}\label{eq: reduced marginal quantile}
     \lefteqn{q_{\tau}(F_Y) = \int \int q_{\tau}(\delta_Y) dF_{Y \mid X} d F_X = c_{1, \tau} \int  P(Y >q_\tau \mid X = x) d F_X(x) + c_{2, \tau}} &&  \label{eq: reduced decomp 1} \\
    && \qquad \qquad \qquad=\int \int q_{\tau}(\delta_Y) dF_{Y \mid X} d F_{(X_{\mathcal S},  \tilde X_{\mathcal S^c})} \nonumber\\
   && \qquad \qquad \qquad =  c_{1, \tau} \int  P\big(Y >q_\tau \mid X = x\big )d F_{(X_{\mathcal S},\tilde X_{\mathcal S^c})} (x_s, \bm 0_{s^c}) + c_{2, \tau}, \label{eq: reduced decomp 2}
\end{eqnarray}
where $c_{1, \tau} = 1/ f_Y(q_\tau)$ and $c_{2, \tau} = q_\tau - c_{1, \tau}(1-\tau)$. The main difference between Eqs.~\ref{eq: reduced decomp 1}, \ref{eq: reduced decomp 2} is the set containing features used in the model. At first glance, it appears to be complicated and trivial to condition on the null variables $\tilde X_{\mathcal S^c}$. However, the additional null variables are inserted to avoid refitting models in the selection procedure, which could be an issue for computationally intensive algorithms such as neural networks and nonparametric models with tensor product interaction terms. 

The two ways of decomposition tell that, under the null hypothesis 
    \begin{equation*}
         H_{0, j}: X_j \in \mathcal{S}^c, 
    \end{equation*}
$q_\tau(F_Y)$ can be estimated using either the initial estimator based on $H$ or an estimator based on $H \backslash \{X_j\}$. An estimator of $\mathcal S$ can be constructed by testing for each feature $X_j$, which is not yet spotted as irrelevant in the initial estimator.

Starting with the estimator $\widehat h$, the marginal quantile estimator, by the decomposition in Eq.~\ref{eq: reduced decomp 1} can be constructed as
\begin{equation}\label{eq: marginal quantile estimator full}
    \hat q_\tau^{\rm full} = \frac{1}{n} \sum_{i = 1}^n \hat c_{1, \tau} \frac{1}{n} \sum_{i' = 1}^n I(\widehat R_{i'} > \hat q_\tau - \widehat h(X_i, \varepsilon_i)) + \hat c_{2, \tau},
\end{equation}
where the estimator $\hat q_\tau$ defined in Eq.~(\ref{eq: quantile estimator}),  the density $f_Y(q_\tau)$ in the constants $c_{1, \tau} = \frac{1}{f_Y(q_\tau)}$ and $c_{2, \tau} = q_\tau - c_{1, \tau}(1-\tau)$ is estimated via kernel estimator in Eq.~(\ref{eq: kernel density extrapolate}). 
The double summations correspond to the double integrals in Eq.~(\ref{eq: reduced decomp 1}) and Eq.~(\ref{eq: reduced decomp 2}). The estimator in Eq.~(\ref{eq: marginal quantile estimator full}) can be assembled in the initialization step in Algorithm~\ref{agthm: full algorithm}. In this initialization step, we test
\begin{equation}\label{eq: goodness of fit}
    H_0^{\rm full}: \widehat h \mbox{ is a good fit}
\mbox{\quad versus \quad }
H_a^{\rm full}: \widehat h \mbox{ is not a good fit},
\end{equation}
i.e., testing if the conditional distribution $F_{Y\mid X}$ is well-approximated by using the estimator $\widehat h(x)$. If the approximation is accurate, the following asymptotic normality hold 
\begin{equation}\label{eq: asymptotic normality}
    \hat q_\tau^{\rm full} -  q_\tau \stackrel{d}{\to} N(0, \int \mbox{IF}(y, q_\tau)^2 dF_Y(y)),
\end{equation}
where $q_\tau$ is estimated using \eqref{eq: quantile estimator} and $\mbox{IF}(y, q_\tau) = \frac{\tau - I\{y \leq q_\tau\}}{f_Y(q_\tau)}$ is binary and can be estimated as 
\begin{eqnarray}\label{eq: hypothesis test asymptotic variance estimator}
   \hat V_n(\tau) = \frac{\tau (1-\tau) }{n f_Y(q_\tau)^2 }. 
\end{eqnarray}
This step gives indication of the ``goodness-of-fit" and is a prerequisite for the pruning step in Algorithm~\ref{agthm: stepwise backward algorithm}. However, a $p$-dimensional vector indicating the importance of features will be reported based on $\hat h$ no matter if it is a good fit or not.

Further, an initial estimate of the features importance $\bm{\hat\beta}^{\rm init}(\tau)$ and the ordered signal strength is denoted as 
$\tilde\beta_{(1)}^{\rm init}(\tau) \leq \ldots \le \tilde\beta_{(p)}^{\rm init}(\tau)$ with features ordered as $(X_{(1)}, \ldots, X_{(p)})$. A candidate set is formulated as 
\begin{equation}\label{eq: candidate irrelevant}
\hat{\mathcal{S}}_0  = \{X_{(j')}: \tilde\beta_j^{\rm init}(\tau) \neq 0 \} 
\end{equation}
with cardinality $|\hat{\mathcal{S}}_0| = s_0$; the ordered candidate set corresponds to $(X_{(p -s_0 + 1)}, \ldots, X_{(p)})$ that is the last $s_0$ elements which are nonzeros in the ordered sequence $(X_{(1)}, \ldots, X_{(p)})$.

The stepwise backward pruning begins with $X_{(p - s_0 + 1)}$ and tests one feature at a time sequentially until the null hypothesis is rejected. In each step $t = j' - (p - s)$, we test 
\begin{equation}\label{eq: stepwise backward}
    H_{0, j' }: X_{(j')} \in S^c \mbox{ versus } H_{a, j'}: X_{(j')} \in S . 
\end{equation}
The test is performed by constructing another the marginal quantile estimator $\hat q^{(-j')}_\tau$ using the set $\{\tilde X_{(1)}, \ldots, \tilde X_{(j')}, X_{(j'+1)},\ldots, X_{(p)}\}$, where $\tilde X_{(j)}, j = 1,\ldots, j'$ have point mass at 0; and the $p$ variables in the set in the original order form a random vector $\tilde X^{(-j')}$.

Similar to Eq.~(\ref{eq: marginal quantile estimator full}), the marginal quantile $q_\tau(F_Y)$ can be estimated using the second decomposition in Eq.~(\ref{eq: reduced decomp 2}) following 
\begin{equation}\label{eq: reduced marginal quantile estimator}
    \hat q_\tau^{(-j')} = \frac{1}{n} \sum_{i = 1}^n \hat c_{1, \tau} \frac{1}{n} \sum_{i' = 1}^n I(R_{i'} > \hat q_\tau - \widehat h(\tilde X_i^{(-j')}, \varepsilon_i)) + \hat c_{2, \tau},
\end{equation}
where $\widehat h(\tilde X^{(-j')}, \varepsilon), $ is the predicted values using $\tilde X^{(-j')}$. Under $H_{0, j'}$, the equality in Eq.~(\ref{eq: reduced decomp 2}) holds, indicating that there isn't a significant change of estimating $q_\tau(F_Y)$ if we replace the ranked $j'$th feature by zeros. Correspondingly, the $(j')$th component of $\hat \beta_{\rm init}(\tau)$ can be truncated to zero.

Algorithm~\ref{agthm: stepwise backward algorithm} summarizes how to form an estimator of the subset of relevant variables $\hat S$ by stepwise backward selection procedure. By using the asymptotic normality in Eq.~(\ref{eq: asymptotic normality}), if $S^c$ contains only the irrelevant features, the marginal quantile estimator $\hat q_\tau^{(-j')}$ follows

\begin{algorithm}[H]
   \label{agthm: multiple quantile algorithm}
    \caption{Stepwise backward pruning via multiple quantiles ${\bm \tau} = (\tau_1, \ldots, \tau_K)^\top$}
     \begin{algorithmic}[1]
              \STATE{
               \textbf{Input:} quantile sequence ${\bm \tau} = (\tau_1, \ldots, \tau_K)^\top$.} 
                \FOR{$k \in \{1, \ldots, K\}$}
                    \STATE{
                    Estimate $\hat\beta^{\rm init}(\tau_k) $ by \eqref{equation:estimator-univariate-output};\\
                    Algorithm~\ref{agthm: full algorithm} with input $\tau_k$ and output $\hat{\mathcal S}_{\tau_k}$, $\hat{\mathcal S}^c_{\tau_k}$.}     
                \ENDFOR
                \STATE{
                 $\hat{\mathcal S}^c = \bigcap_k \mathcal S^c_{\tau_k}$; $\hat{\mathcal S} = H \backslash \hat{\mathcal S}^c$;\\
                 $\hat \beta_j^{\rm init}(\tau_k) \leftarrow 0, j \in I = \{j: X_j \in \hat{\mathcal S}^c\}, k \in \{1, \ldots, K\}$.\\
                 
                 \RETURN{$\hat \beta(\tau_k) \leftarrow \hat \beta^{\rm init}(\tau_k), k = 1, \ldots, K$.}
                }
            \end{algorithmic}
          \end{algorithm}

\begin{algorithm}[H]
   \label{agthm: full algorithm}
    \caption{Estimate the subset $\mathcal S$ at quantile level $\tau$}
     \begin{algorithmic}[1]
              \STATE{
              \textbf{Input:} quantile level $\tau$.\\
              \textbf{Output:} estimated subsets $\hat S$ and $\hat S^c$.\\
              Estimate $q_{\tau}(F_Y) = \int \int q_{\tau}(\delta_Y) dF_{Y \mid X} d F_X $ by the estimator in Eq.~(\ref{eq: marginal quantile estimator full}). \\
              Test the hypothesis in Eq.~(\ref{eq: goodness of fit}) by
               \[
                T_{\rm full, \tau} = \sqrt{\frac{1}{\hat V_{\tau, n}}} (\hat q_{\tau}^{\rm full} - \hat q_{\tau}).
                \]
                }
                \IF{$P_{\rm full, \tau} = 2(1 - \Phi(|T_{\rm full, \tau}|)) > \alpha $} \STATE{Algorithm~\ref{agthm: stepwise backward algorithm} backward elimination\\
                \RETURN{$\hat{\mathcal S}_\tau$; $\hat{\mathcal S}_\tau^c$.}}
                \ELSE 
                \STATE{$\hat{\mathcal S}_\tau \leftarrow H$; $\hat{\mathcal S}_\tau^c \leftarrow \varnothing$.\\
                \RETURN{$\hat{\mathcal S}_\tau$; $\hat{\mathcal S}_\tau^c$.}}
                \ENDIF
           \end{algorithmic}
          \end{algorithm}

\begin{algorithm}[H]
   \label{agthm: stepwise backward algorithm}
    \caption{Stepwise backward pruning at quantile level $\tau$ and significance level $\alpha$}
            \begin{algorithmic}[1]
              \STATE{\textbf{Initialization:}
              $\hat{\mathcal S}$ by Eq.~(\ref{eq: candidate irrelevant}); $\hat{\mathcal S}^c = H \backslash \hat{\mathcal S}$. 
              }
            \FOR{$j' \in \{p - s_0 + 1, \ldots, p\}$}
               \STATE{
                Let $\hat{\mathcal S}_{ j'} = \hat{\mathcal S} \backslash \{ X_{(j')} \}$.\\
                Estimate \[q_{\tau} (F_Y) = \int \int q_{\tau}(\delta_Y) dF_{Y \mid (X_{\hat{\mathcal S}_{j'}}, \tilde X_{\mathcal S_{j'}^c})} d F_{(X_{\mathcal S_{j'}}, \tilde X_{\mathcal S_{j'}^c})},\] 
                by Eq.~(\ref{eq: reduced marginal quantile estimator}) and calculate test statistic
                \[
                T_{\tau, j'} = \sqrt{\frac{1}{\hat V_{\tau, n}}} (\hat q_{\tau}^{(-j')}  - \hat q_{\tau}).
                \]
                \IF{$P_{j'} = 2(1 - \Phi(|T_{( j)}|)) > \alpha $} 
                        \STATE{ $\hat{\mathcal S} \leftarrow \hat{\mathcal S} \backslash \{X_{(j')} \}; 
                        T \leftarrow j'$.}
                \ELSE \STATE{break}
                \ENDIF
                }
            \ENDFOR
            \RETURN{$\hat{\mathcal S}_\tau \leftarrow \hat{\mathcal S}$; $\hat{\mathcal S}_\tau^c \leftarrow H \backslash \hat{\mathcal S}_\tau$.
            }
           \end{algorithmic}
          \end{algorithm}


\section{Numerical Performance}

\subsection{Synthetic Experiments}

We consider testing the proposed estimator on different nonlinear models. To evaluate the performance, one can choose one variable having only the linear term in the data generating process. Further, the following should be expected: (1) the importance measures of the linear terms should be a constant across the quantile levels and should be approximately equal to the coefficient in the data generating process; (2) the irrelevant features should have importance measures being zero; (3) the measures for the nonlinear terms vary at different quantile levels. The quantile levels considered are $\tau = 0.1, 0.3, 0.5, 0.7, 0.9$.

The features are generated from a multivariate Gaussian distribution $X \sim N(0, \Sigma)$, where $\Sigma_{jj'} = 0.5^{|j-j'|}, j, j' = 1, \ldots, p$ using seed number 5. The error distributions considered are $N(0,1)$, $t_3$, $\mbox{Exp}(2)$, and $\mbox{Cauchy}(0, 1)$; they are generated using seed numbers 1 to 500 for $R = 500$ times replication. In the low dimensional setting, we consider sample size $n =1000$ and the number of features $p = 4$. In the high dimensional setting, the same sample size $n=1000$ is considered and the number of features is set to be $p = 500$.

\subsubsection{Low-dimensional Settings}\label{ssec: low dim settings}
We consider the following six data generating processes where $X_2$ is the linear term with true coefficient -5. In the first three data generating processes, $X_1, X_2$ enter the true model and different nonlinear functions are applied to $X_1$. In data generating processes 4-6, $X_3$ interacts with the nonlinear functions of $X_1$, $X_4$ is irrelevant to $Y$. In the last three, we mainly want to allow heteroscedasticity, that is, $\varepsilon$ interacts with some of the features.
\begin{enumerate}
    \item (Model 1) $Y = (1 + 2 X_1)^2 -5 X_2 + \varepsilon$;
    \item (Model 2)  
    $Y = 1 +  \exp(2X_1) I(|X_1]| \leq 1) -5 X_2 + \varepsilon$;
    \item (Model 3) $Y = 1 + 2\cos(X_1) - 5X_2 + \varepsilon$;
    \item (Model 4) $Y = (1 + 2X_1 + X_3)^2 -5 X_2 + \varepsilon$;
   \item (Model 5) $Y = 1 +  \exp(2 X_1 +X_3)I(|X_1|\leq 1) -5X_2 +\varepsilon$;
    \item (Model 6) $Y = (1 + 2\cos(X_1) + X_3 )^2 -5 X_2 + \varepsilon$;
    \item (Model 7) $Y = (1 + 2X_1 + X_3)^2 -5 X_2 + \exp(X_1) \varepsilon$;
    \item (Model 8) $Y = 1 +  \exp(2 X_1 +X_3)I(|X_1|\leq 1) -5X_2 + \exp(X_1) \varepsilon$;
    \item (Model 9) $Y = 1 + 2\cos(X_1) - 5X_2 + \exp(X_1) \varepsilon$.
\end{enumerate}


For estimating $h(x)$, we consider the generalized additive models implemented in the R package $\texttt{mgcv}$. For the models~1, 2, 3, we add spline terms of all fours features; for models~4, 5, 6, we add a tensor product smooth term between $X_1$ and $X_3$. Models~7, 8, 9 are heteroscedastic compared to models 4, 5, 3. The empirical means and standard deviations of the $R = 500$ simulation replications are reported in Table~\ref{table: gam fit summary models 1-3}, \ref{table: gam fit summary model 4-6}, \ref{table: gam fit summary models 7-9}. For the backward elimination, we recommend checking if both the mean and the standard deviation are equal to zero. In certain cases, the standard deviations are nonzero, which suggests that the trimming steps did not find all the irrelevant features in the 500 replications. However, the standard deviations should be close to zero to indicate a relatively good performance. Under heteroscedasticity, the feature importance estimates are close to those in the the corresponding homoscedastic settings. For prunning, we observe that, although the averaged estimates of irrelevant feature importance are still zero or near-zero, the standard deviations increase compared to those in the corresponding homoscedastic settings.

\begin{table}[!htb]
	\centering
    \caption{The empirical means and standard deviations of $\hat\beta(\tau)$ for models 1-3 in the low dimensional settings. The estimator $\hat h(x)$ is a GAM fit.}
	\bgroup
	\def\arraystretch{1.2}
	\begin{adjustbox}{max width=\textwidth}
	\begin{tabular}{c |c| c| c c c c| c c c c |c c c c| c ccc|cccc| } \hline
  \multicolumn{23}{c}{Model 1}\\ \hline
 & &$\tau$ & \multicolumn{4}{c}{$0.1$} & \multicolumn{4}{c}{$0.3$} &
 \multicolumn{4}{c}{$0.5$} & \multicolumn{4}{c}{$0.7$} & \multicolumn{4}{c}{$0.9$}\\ \hline 
 && & $\beta_1(\tau)$ & $\beta_2(\tau)$ & $\beta_3(\tau)$ &  $\beta_4(\tau)$ &$\beta_1(\tau)$ & $\beta_2(\tau)$ & $\beta_3(\tau)$ &  $\beta_4(\tau)$&$\beta_1(\tau)$ & $\beta_2(\tau)$ & $\beta_3(\tau)$ &  $\beta_4(\tau)$&$\beta_1(\tau)$ & $\beta_2(\tau)$ & $\beta_3(\tau)$ &  $\beta_4(\tau)$ & $\beta_1(\tau)$ & $\beta_2(\tau)$ & $\beta_3(\tau)$ &  $\beta_4(\tau)$ \\ \hline
 \multirow{8}{*}{$\varepsilon$}& \multirow{2}{*}{$N(0,1)$} & $\bar\beta$ & 4.37 & -5.02 & 0.00 & 0.00 &  4.40  & -5.05 & 0.00 & 0.00 & 4.41 & -5.07 &0.00 & 0.00 &  4.63 & -5.31 & 0.00 & 0.00 &  4.20  &-4.83 & 0.00 & 0.00 \\
&&$\sigma(\beta)$ & 0.21 & 0.26 & 0.00 &0.00 & 0.13 & 0.16 &0.00 &0.00 & 0.17 &0.25 & 0.00 & 0.00 & 0.15 & 0.16 & 0.00 & 0.00 & 0.15 & 0.18 & 0.00 & 0.00 \\ 
& \multirow{2}{*}{$t_3$} & $\bar\beta$ & 4.44 & -5.10 & 0.00 & 0.00 & 4.58 & -5.26 & 0.00 & 0.00 &  4.37 & -5.02 & 0.00 & 0.00 & 4.65 & -5.34 & 0.00 &0.00 &  4.21 & -4.84 & 0.00 &  0.00 \\
&&$\sigma(\beta)$ & 0.21 & 0.25 & 0.00 & 0.00 & 0.14 & 0.16 & 0.00 & 0.00 & 0.10 & 0.11 & 0.00 & 0.00 & 0.11 & 0.13 & 0.00 & 0.00 & 0.13 & 0.15 & 0.00 & 0.00 \\
& \multirow{2}{*}{$\mbox{Exp}(2)$} & $\bar\beta$ & 4.24 & -4.87 & 0.00 & 0.00 & 4.49 & -5.16 & 0.00 & 0.00 & 4.41 & -5.07 & 0.00 & 0.00 & 4.59 & -5.28 & 0.00 & 0.00 & 4.25 & -4.88 & 0.00 & 0.00 \\
&&$\sigma(\beta)$ & 0.24 & 0.27 & 0.00& 0.00 & 0.15 & 0.18 & 0.00 & 0.00& 0.14 & 0.16 & 0.00 & 0.00 & 0.12 & 0.15 & 0.00 & 0.00 & 0.15 & 0.18 & 0.00 & 0.00 \\ 
& \multirow{2}{*}{$\mbox{Cauchy}(0,1)$} & $\bar\beta$ & 5.07 & -5.83 & 0.00 & 0.00 & 5.62 & -6.46 & 0.00 & 0.00 & 4.36 & -5.01 & 0.00 & 0.00 & 4.81 & -5.53 & 0.00 & 0.00 & 5.19 & -5.96 & 0.00 & 0.00 \\
&&$\sigma(\beta)$ & 0.92 & 1.06 & 0.00 & 0.00 & 0.65 & 0.74 & 0.00 & 0.00 & 0.42 & 0.49 & 0.00 & 0.00 & 0.34 & 0.40 & 0.00 & 0.00 & 0.77 & 0.89 & 0.00 & 0.00  \\ \hline 
  \multicolumn{23}{c}{Model 2}\\ \hline
 & &$\tau$ & \multicolumn{4}{c}{$0.1$} & \multicolumn{4}{c}{$0.3$} &
 \multicolumn{4}{c}{$0.5$} & \multicolumn{4}{c}{$0.7$} & \multicolumn{4}{c}{$0.9$}\\ \hline 
 && & $\beta_1(\tau)$ & $\beta_2(\tau)$ & $\beta_3(\tau)$ &  $\beta_4(\tau)$ &$\beta_1(\tau)$ & $\beta_2(\tau)$ & $\beta_3(\tau)$ &  $\beta_4(\tau)$&$\beta_1(\tau)$ & $\beta_2(\tau)$ & $\beta_3(\tau)$ &  $\beta_4(\tau)$&$\beta_1(\tau)$ & $\beta_2(\tau)$ & $\beta_3(\tau)$ &  $\beta_4(\tau)$ & $\beta_1(\tau)$ & $\beta_2(\tau)$ & $\beta_3(\tau)$ &  $\beta_4(\tau)$ \\ \hline
 \multirow{8}{*}{$\varepsilon$}& \multirow{2}{*}{$N(0,1)$} & $\bar\beta$ &  0.01 & -4.95 & 0.00 & 0.00 & 0.16 & -5.11 & 0.00 & 0.00 & 0.53 & -5.25 &  0.00 & 0.00 & 0.51 & -4.95 & 0.00 & 0.00 & 0.09 & -4.97 & 0.00 & 0.00 
 \\
&&$\sigma(\beta)$ & 0.06 & 0.21 & 0.00 & 0.00 & 0.27 & 0.15 & 0.02 & 0.00 & 0.22 & 0.15 & 0.02 & 0.01 & 0.20 & 0.16 & 0.02 & 0.01 & 0.21 & 0.22 & 0.00 & 0.00 
 \\ 
& \multirow{2}{*}{$t_3$} & $\bar\beta$ & 0.00 & -5.02 & 0.00 & 0.00 &  0.13 & -5.10 & 0.00 & 0.00 & 0.52 & -5.22 & 0.00 & 0.00 & 0.52 & -4.91 & 0.00 &  0.00 & 0.07 & -4.96 & 0.00 & 0.00 
\\
&&$\sigma(\beta)$ & 0.03 & 0.19 & 0.00 & 0.00 & 0.25 & 0.13 & 0.00 & 0.00 & 0.23 & 0.14 & 0.00 & 0.00 & 0.18 & 0.13 & 0.00 & 0.00 & 0.19 & 0.20 & 0.00 & 0.00 \\
& \multirow{2}{*}{$\mbox{Exp}(2)$} & $\bar\beta$ & 0.00 & -4.96 & 0.00 & 0.00 & 0.11 & -5.18 & 0.00 & 0.00 & 0.6 & -5.4 & 0.0 & 0.0 & 0.54 & -4.96 &  0.00 & 0.00 & 0.07 & -4.98 & 0.00 & 0.00 \\
&&$\sigma(\beta)$ & 0.03 & 0.20 & 0.00 & 0.00 & 0.24 & 0.15 & 0.00 & 0.00 & 0.16 & 0.15 & 0.00 & 0.00 & 0.16 & 0.14 & 0.00 & 0.00 & 0.20 & 0.19 & 0.00 & 0.00 \\ 
& \multirow{2}{*}{$\mbox{Cauchy}(0,1)$} & $\bar\beta$ & 0.00 & -5.07 & 0.00 & 0.00 & 0.11 & -5.12 & 0.00 & 0.00 & 0.57 & -5.02 & 0.00 & 0.00 & 0.55 & -4.85 & 0.00 & 0.00 & 0.02 & -4.65 & 0.00 & 0.00 \\ 
&& $\sigma(\beta)$ & 0.00 & 0.18 & 0.00 & 0.00 & 0.24 & 0.11 & 0.00 & 0.00 & 0.12 & 0.14 & 0.00 & 0.00 & 0.12 & 0.13 & 0.00 & 0.00 & 0.11 & 0.24 & 0.00 & 0.00  \\ \hline 
  \multicolumn{23}{c}{Model 3}\\ \hline
 & &$\tau$ & \multicolumn{4}{c}{$0.1$} & \multicolumn{4}{c}{$0.3$} &
 \multicolumn{4}{c}{$0.5$} & \multicolumn{4}{c}{$0.7$} & \multicolumn{4}{c}{$0.9$}\\ \hline 
 && & $\beta_1(\tau)$ & $\beta_2(\tau)$ & $\beta_3(\tau)$ &  $\beta_4(\tau)$ &$\beta_1(\tau)$ & $\beta_2(\tau)$ & $\beta_3(\tau)$ &  $\beta_4(\tau)$&$\beta_1(\tau)$ & $\beta_2(\tau)$ & $\beta_3(\tau)$ &  $\beta_4(\tau)$&$\beta_1(\tau)$ & $\beta_2(\tau)$ & $\beta_3(\tau)$ &  $\beta_4(\tau)$ & $\beta_1(\tau)$ & $\beta_2(\tau)$ & $\beta_3(\tau)$ &  $\beta_4(\tau)$ \\ \hline
 \multirow{8}{*}{$\varepsilon$}& \multirow{2}{*}{$N(0,1)$} & $\bar\beta$ &  -0.03 & -5.00 & 0.00 & 0.00 & -0.03 & -5.04 & 0.00 & 0.00 & -0.03 & -5.19 & 0.00 & 0.00 & -0.03 & -5.08 & 0.00 & 0.00 & -0.03 & -4.83 & 0.00 & 0.00 
 \\
&&$\sigma(\beta)$ & 0.06 & 0.54 & 0.06 & 0.03 & 0.04 & 0.21 & 0.03 & 0.03 & 0.04 & 0.14 & 0.04 & 0.03 & 0.05 & 0.30 & 0.05 & 0.03 & 0.07 & 0.87 & 0.08 & 0.03 \\ 
& \multirow{2}{*}{$t_3$} & $\bar\beta$ & -0.03 & -5.14 & 0.00 & 0.00 & -0.03 & -5.00 & 0.00 & 0.00 & -0.03 & -5.25 & 0.00 & 0.00 & -0.03 & -5.07 & 0.00 & 0.00 & -0.03 & -4.66 & 0.00 & 0.00 \\
&&$\sigma(\beta)$ & 0.04 & 0.19 & 0.04 & 0.03 & 0.04 & 0.14 & 0.04 & 0.03 & 0.04 & 0.13 & 0.04 & 0.03 & 0.04 & 0.14 & 0.04 & 0.03 & 0.03 & 0.24 & 0.03 & 0.03 \\
& \multirow{2}{*}{$\mbox{Exp}(2)$} & $\bar\beta$ & -0.03 & -5.01 & 0.00 & 0.00 & -0.03 & -5.10 & 0.00 & 0.00 & -0.03 & -4.99 & 0.00 & 0.00 & -0.03 & -4.81 & 0.00 & 0.00 & -0.03 & -5.17 & 0.00 & 0.00 
\\
&&$\sigma(\beta)$ & 0.04 & 0.21 & 0.03 & 0.03 & 0.04 & 0.16 & 0.03 & 0.03 & 0.04 & 0.15 & 0.03 & 0.03 & 0.03 & 0.14 & 0.03 & 0.03 & 0.04 & 0.22 & 0.03 & 0.03 \\ 
& \multirow{2}{*}{$\mbox{Cauchy}(0,1)$} & $\bar\beta$ & -0.04 & -5.84 & 0.00 & 0.00 & -0.03 & -5.03 & 0.00 & 0.00 & -0.05 & -6.25 & 0.00 & 0.00 &
-0.04 & -5.78 & 0.00 & 0.00 & -0.03 & -4.49 & 0.00 & 0.00 \\ 
&& $\sigma(\beta)$ & 0.04 & 0.64 & 0.04 & 0.03 & 0.04 & 0.50 & 0.04 & 0.03 & 0.05 & 0.50 & 0.04 & 0.04 & 0.04 & 0.70 & 0.04 & 0.03 & 0.03 & 0.95 & 0.03 & 0.03 \\ \hline 
\end{tabular}\label{table: gam fit summary models 1-3}
     \end{adjustbox}
    \egroup
 \end{table}

\begin{table}[!htb]
	\centering
 \caption{The empirical means and standard deviations of $\hat\beta(\tau)$ for models 4-6 in the low dimensional settings. The estimator $\hat h(x)$ is a GAM fit.}
	\bgroup
	\def\arraystretch{1.2}
	\begin{adjustbox}{max width=\textwidth}
	\begin{tabular}{c |c| c| c c c c| c c c c |c c c c| c ccc|cccc| } \hline
  \multicolumn{23}{c}{Model 4}\\ \hline
 & &$\tau$ & \multicolumn{4}{c}{$0.1$} & \multicolumn{4}{c}{$0.3$} &
 \multicolumn{4}{c}{$0.5$} & \multicolumn{4}{c}{$0.7$} & \multicolumn{4}{c}{$0.9$}\\ \hline 
 && & $\beta_1(\tau)$ & $\beta_2(\tau)$ & $\beta_3(\tau)$ &  $\beta_4(\tau)$ &$\beta_1(\tau)$ & $\beta_2(\tau)$ & $\beta_3(\tau)$ &  $\beta_4(\tau)$&$\beta_1(\tau)$ & $\beta_2(\tau)$ & $\beta_3(\tau)$ &  $\beta_4(\tau)$&$\beta_1(\tau)$ & $\beta_2(\tau)$ & $\beta_3(\tau)$ &  $\beta_4(\tau)$ & $\beta_1(\tau)$ & $\beta_2(\tau)$ & $\beta_3(\tau)$ &  $\beta_4(\tau)$ \\ \hline
 \multirow{8}{*}{$\varepsilon$}& \multirow{2}{*}{$N(0,1)$} & $\bar\beta$ & 
  4.01 & -4.73 & 2.00 & 0.00 & 4.12 & -4.86 & 2.05 & 0.00 & 4.42 & -5.21 & 2.20 & 0.00 & 4.52 & -5.34 & 2.25 & 0.00 & 4.33 & -5.11 & 2.02 & 0.00 \\
  &&$\sigma(\beta)$ & 0.22 & 0.26 & 0.11 & 0.00 & 0.12 & 0.15 & 0.07 & 0.00 & 0.24 & 0.36 & 0.10 & 0.00 & 0.13 & 0.19 & 0.07 & 0.00 & 0.67 & 0.94 & 0.60 & 0.00   \\ 
& \multirow{2}{*}{$t_3$} & $\bar\beta$ & 4.04 & -4.77 & 2.01 & 0.00 & 4.19 & -4.94 & 2.09 & 0.00 & 4.38 & -5.17 & 2.18 & 0.00 &  4.61 & -5.43 & 2.29 & 0.00 & 4.29 & -5.06 & 1.90 & 0.00 
\\
&&$\sigma(\beta)$ & 0.26 & 0.31 & 0.14 & 0.00 & 0.12 & 0.14 & 0.07 & 0.00 & 
0.10 & 0.12 & 0.06 & 0.00 & 0.10 & 0.11 & 0.06 & 0.00 & 0.13 & 0.15 & 0.67 & 0.00 \\
& \multirow{2}{*}{$\mbox{Exp}(2)$} & $\bar\beta$ & 4.19 & -4.95 & 2.09 & 0.00 & 4.17 & -4.92 & 2.07 & 0.00 & 4.56 & -5.38 & 2.27 & 0.00 & 4.54 & -5.36 & 2.26 & 0.00 & 4.14 & -4.88 & 1.94 & 0.00 
\\
&&$\sigma(\beta)$ & 0.29 & 0.34 & 0.15 & 0.00 & 0.14 & 0.16 & 0.08 & 0.00 & 0.11 & 0.14 & 0.08 & 0.00 & 0.11 & 0.13 & 0.07 & 0.00 & 0.15 & 0.18 & 0.50 & 0.00 \\ 
& \multirow{2}{*}{$\mbox{Cauchy}(0,1)$} & $\bar\beta$ &  5.31 & -6.27 & 2.64 & 0.00 & 4.68 & -5.53 & 2.33 & 0.00 & 3.80 & -4.49 & 1.89 & 0.00 &  5.11 & -6.03 & 2.54 & 0.00 & 3.91 & -4.62 & 1.74 & 0.00 \\
&&$\sigma(\beta)$ & 1.33 & 1.57 & 0.66 & 0.00 & 0.53 & 0.63 & 0.26 & 0.00 & 0.47 & 0.56 & 0.23 & 0.00 & 0.38 & 0.46 & 0.18 & 0.00 & 0.43 & 0.51 & 0.62 & 0.00 
\\ \hline 
  \multicolumn{23}{c}{Model 5}\\ \hline
 & &$\tau$ & \multicolumn{4}{c}{$0.1$} & \multicolumn{4}{c}{$0.3$} &
 \multicolumn{4}{c}{$0.5$} & \multicolumn{4}{c}{$0.7$} & \multicolumn{4}{c}{$0.9$}\\ \hline 
 && & $\beta_1(\tau)$ & $\beta_2(\tau)$ & $\beta_3(\tau)$ &  $\beta_4(\tau)$ &$\beta_1(\tau)$ & $\beta_2(\tau)$ & $\beta_3(\tau)$ &  $\beta_4(\tau)$&$\beta_1(\tau)$ & $\beta_2(\tau)$ & $\beta_3(\tau)$ &  $\beta_4(\tau)$&$\beta_1(\tau)$ & $\beta_2(\tau)$ & $\beta_3(\tau)$ &  $\beta_4(\tau)$ & $\beta_1(\tau)$ & $\beta_2(\tau)$ & $\beta_3(\tau)$ &  $\beta_4(\tau)$ \\ \hline
 \multirow{8}{*}{$\varepsilon$}& \multirow{2}{*}{$N(0,1)$} & $\bar\beta$ &   0.50 & -5.57 & 2.09 & 0.08 & 0.33 & -4.92 & 1.85 & 0.00 & 0.53 & -4.70 & 1.76 & 0.00 & 0.52 & -4.60 & 1.73 & 0.00 & 0.57 & -5.02 & 1.88 & 0.10 
 \\
&&$\sigma(\beta)$ & 0.26 & 0.26 & 0.11 & 0.06 & 0.28 & 0.16 & 0.07 & 0.02 & 0.05 & 0.15 & 0.07 & 0.00 & 0.04 & 0.14 & 0.06 & 0.01 & 0.05 & 0.22 & 0.09 & 0.04  \\ 
& \multirow{2}{*}{$t_3$} & $\bar\beta$ & 0.52 & -5.59 & 2.10 & 0.09 & 0.32 & -4.95 & 1.86 & 0.00 & 0.53 & -4.68 & 1.76 & 0.00 & 0.52 & -4.62 & 1.74 & 0.00 & 0.58 & -5.06 & 1.90 & 0.10 
\\
&&$\sigma(\beta)$ & 0.25 & 0.25 & 0.10 & 0.05 & 0.28 & 0.14 & 0.07 & 0.02 & 0.05 & 0.15 & 0.07 & 0.00 & 0.05 & 0.12 & 0.06 & 0.00 & 0.05 & 0.21 & 0.09 & 0.04\\
& \multirow{2}{*}{$\mbox{Exp}(2)$} & $\bar\beta$ & 0.50 & -5.61 & 2.10 & 0.09 & 0.33 & -4.96 & 1.86 & 0.01 & 0.53 & -4.67 & 1.75 & 0.00 & 0.52 & -4.65 & 1.74 & 0.00 & 0.58 & -5.07 & 1.90 & 0.10 \\
&&$\sigma(\beta)$ & 0.26 & 0.26 & 0.10 & 0.06 & 0.28 & 0.15 & 0.07 & 0.02 & 0.04 & 0.15 & 0.07 & 0.00 & 0.08 & 0.12 & 0.05 & 0.01 & 0.05 & 0.22 & 0.09 & 0.04 
\\ 
& \multirow{2}{*}{$\mbox{Cauchy}(0,1)$} & $\bar\beta$ & 0.62 & -5.60 & 2.10 & 0.11 & 0.35 & -4.99 & 1.87 & 0.00 & 0.53 & -4.65 & 1.75 & 0.00 & 0.53 & -4.68 & 1.76 & 0.00 & 0.58 & -5.14 & 1.93 & 0.10 \\ 
&& $\sigma(\beta)$ & 0.09 & 0.14 & 0.07 & 0.04 & 0.28 & 0.09 & 0.06 & 0.01 & 0.04 & 0.13 & 0.07 & 0.00 & 0.05 & 0.07 & 0.05 & 0.01 & 0.05 & 0.19 & 0.08 & 0.04 \\ \hline 
  \multicolumn{23}{c}{Model 6}\\ \hline
 & &$\tau$ & \multicolumn{4}{c}{$0.1$} & \multicolumn{4}{c}{$0.3$} &
 \multicolumn{4}{c}{$0.5$} & \multicolumn{4}{c}{$0.7$} & \multicolumn{4}{c}{$0.9$}\\ \hline 
 && & $\beta_1(\tau)$ & $\beta_2(\tau)$ & $\beta_3(\tau)$ &  $\beta_4(\tau)$ &$\beta_1(\tau)$ & $\beta_2(\tau)$ & $\beta_3(\tau)$ &  $\beta_4(\tau)$&$\beta_1(\tau)$ & $\beta_2(\tau)$ & $\beta_3(\tau)$ &  $\beta_4(\tau)$&$\beta_1(\tau)$ & $\beta_2(\tau)$ & $\beta_3(\tau)$ &  $\beta_4(\tau)$ & $\beta_1(\tau)$ & $\beta_2(\tau)$ & $\beta_3(\tau)$ &  $\beta_4(\tau)$ \\ \hline
 \multirow{8}{*}{$\varepsilon$}& \multirow{2}{*}{$N(0,1)$} & $\bar\beta$ &  -0.59 & -4.82 & 4.22 & 0.00 & -0.62 & -5.03 & 4.41 & 0.00 & -0.63 & -5.14 & 4.50 & 0.00 & -0.64 & -5.20 & 4.56 & 0.00 & -0.58 & -4.72 & 4.14 & 0.00  \\
&&$\sigma(\beta)$ & 0.04 & 0.21 & 0.19 & 0.00 & 0.05 & 0.17 & 0.13 & 0.00 & 0.04 & 0.13 & 0.12 & 0.00 & 0.04 & 0.14 & 0.13 & 0.00 & 0.06 & 0.29 & 0.21 & 0.00 \\ 
& \multirow{2}{*}{$t_3$} & $\bar\beta$ & -0.59 & -4.79 & 4.20 & 0.00 & -0.61 & -5.02 & 4.40 & 0.00 & -0.63 & -5.16 & 4.52 & 0.00 & -0.65 & -5.32 & 4.66 & 0.00 & -0.57 & -4.69 & 4.12 & 0.00 
 \\
&&$\sigma(\beta)$ & 0.05 & 0.21 & 0.19 & 0.00 & 0.04 & 0.14 & 0.13 & 0.00 & 0.04 & 0.12 & 0.11 & 0.00 & 0.05 & 0.15 & 0.14 & 0.00 & 0.04 & 0.20 & 0.18 & 0.00 \\
& \multirow{2}{*}{$\mbox{Exp}(2)$} & $\bar\beta$ & -0.62 & -5.06 & 4.43 & 0.00 & -0.60 & -4.92 & 4.31 & 0.00 & -0.63 & -5.15 & 4.51 & 0.00 & -0.65 & -5.32 & 4.66 & 0.00 & -0.54 & -4.45 & 3.90 & 0.00 \\
&&$\sigma(\beta)$ & 0.05 & 0.25 & 0.22 & 0.00 & 0.04 & 0.16 & 0.14 & 0.00 & 0.04 & 0.14 & 0.13 & 0.00 & 0.04 & 0.15 & 0.13 & 0.00 & 0.04 & 0.22 & 0.19 &0.00  \\ 
& \multirow{2}{*}{$\mbox{Cauchy}(0,1)$} & $\bar\beta$ & -0.65 & -5.25 & 4.61 & 0.00 & -0.56 & -4.59 & 4.02 & 0.00 & -0.64 & -5.19 & 4.55 & 0.00 & -0.77 & -6.30 & 5.52 & 0.00 & -0.51 & -4.16 & 3.64 & 0.00 
 \\ 
&& $\sigma(\beta)$ & 0.08 & 0.63 & 0.56 & 0.00 & 0.06 & 0.42 & 0.37 & 0.00 & 0.06 & 0.32 & 0.29 & 0.00 & 0.07 & 0.52 & 0.47 & 0.00 & 0.09 & 0.67 & 0.58 & 0.00 
 \\ \hline 
\end{tabular}\label{table: gam fit summary model 4-6}
     \end{adjustbox}
    \egroup
 \end{table}

\begin{table}[!htb]
	\centering
    \caption{The empirical means and standard deviations of $\hat\beta(\tau)$ for models 1-3 in the low dimensional settings. The estimator $\hat h(x)$ is a GAM fit.}
	\bgroup
	\def\arraystretch{1.2}
	\begin{adjustbox}{max width=\textwidth}
	\begin{tabular}{c |c| c| c c c c| c c c c |c c c c| c ccc|cccc| } \hline
  \multicolumn{23}{c}{Model 7}\\ \hline
 & &$\tau$ & \multicolumn{4}{c}{$0.1$} & \multicolumn{4}{c}{$0.3$} &
 \multicolumn{4}{c}{$0.5$} & \multicolumn{4}{c}{$0.7$} & \multicolumn{4}{c}{$0.9$}\\ \hline 
 && & $\beta_1(\tau)$ & $\beta_2(\tau)$ & $\beta_3(\tau)$ &  $\beta_4(\tau)$ &$\beta_1(\tau)$ & $\beta_2(\tau)$ & $\beta_3(\tau)$ &  $\beta_4(\tau)$&$\beta_1(\tau)$ & $\beta_2(\tau)$ & $\beta_3(\tau)$ &  $\beta_4(\tau)$&$\beta_1(\tau)$ & $\beta_2(\tau)$ & $\beta_3(\tau)$ &  $\beta_4(\tau)$ & $\beta_1(\tau)$ & $\beta_2(\tau)$ & $\beta_3(\tau)$ &  $\beta_4(\tau)$ \\ \hline
 \multirow{8}{*}{$\varepsilon$}& \multirow{2}{*}{$N(0,1)$} & $\bar\beta$ &  4.14 & -4.88 & 2.07 & 0.00 & 4.22 & -4.98 & 2.10 & 0.00 & 4.24 & -5.00 & 2.07 & 0.00 & 4.30 & -5.08 & 2.14 & 0.00 & 4.10 & -4.83 & 2.04 & 0.00 
\\
&&$\sigma(\beta)$ & 0.45 & 0.48 & 0.23 & 0.00 & 0.26 & 0.23 & 0.18 & 0.00 & 0.26 & 0.21 & 0.32 & 0.00 & 0.24 & 0.20 & 0.20 & 0.00 & 0.28 & 0.22 & 0.18 & 0.02 
\\ 
& \multirow{2}{*}{$t_3$} & $\bar\beta$ & 4.13 & -4.86 & 2.05 & 0.00 & 4.24 & -4.99 & 2.11 & 0.00 & 4.28 & -5.04 & 2.11 & 0.00 & 4.41 & -5.19 & 2.17 & 0.00 & 4.18 & -4.92 & 1.98 & 0.00 
\\
&&$\sigma(\beta)$ & 0.47 & 0.51 & 0.24 & 0.00 & 0.28 & 0.24 & 0.14 & 0.00 & 0.28 & 0.21 & 0.23 & 0.01 & 0.27 & 0.21 & 0.23 & 0.00 & 0.31 & 0.24 & 0.45 & 0.01 
 \\
& \multirow{2}{*}{$\mbox{Exp}(2)$} & $\bar\beta$ & 4.16 & -4.88 & 2.05 & 0.00 &  4.19 & -4.93 & 2.07 & 0.00 & 4.24 & -4.98 & 2.09 & 0.00 & 4.37 & -5.14 & 2.15 & 0.00 &  4.18 & -4.91 & 2.06 & 0.00 
\\
&&$\sigma(\beta)$ & 0.47 & 0.50 & 0.23 & 0.00 & 0.25 & 0.25 & 0.17 & 0.00 & 0.25 & 0.19 & 0.13 & 0.00 & 0.23 & 0.22 & 0.15 & 0.00 & 0.31 & 0.24 & 0.13 & 0.04 
 \\ 
& \multirow{2}{*}{$\mbox{Cauchy}(0,1)$} & $\bar\beta$ & 4.59 & -5.41 & 2.27 & 0.00 & 4.65 & -5.48 & 2.31 & 0.00 & 3.99 & -4.71 & 1.98 & 0.00 & 4.88 & -5.76 & 2.42 & 0.00 & 4.07 & -4.81 & 1.70 & 0.00 
\\
&&$\sigma(\beta)$ & 1.29 & 1.50 & 0.63 & 0.02 & 0.54 & 0.63 & 0.29 & 0.01 & 0.53 & 0.60 & 0.27 & 0.00 & 0.47 & 0.53 & 0.25 & 0.01 & 0.49 & 0.53 & 0.77 & 0.03 
 \\ \hline 
  \multicolumn{23}{c}{Model 8}\\ \hline
 & &$\tau$ & \multicolumn{4}{c}{$0.1$} & \multicolumn{4}{c}{$0.3$} &
 \multicolumn{4}{c}{$0.5$} & \multicolumn{4}{c}{$0.7$} & \multicolumn{4}{c}{$0.9$}\\ \hline 
 && & $\beta_1(\tau)$ & $\beta_2(\tau)$ & $\beta_3(\tau)$ &  $\beta_4(\tau)$ &$\beta_1(\tau)$ & $\beta_2(\tau)$ & $\beta_3(\tau)$ &  $\beta_4(\tau)$&$\beta_1(\tau)$ & $\beta_2(\tau)$ & $\beta_3(\tau)$ &  $\beta_4(\tau)$&$\beta_1(\tau)$ & $\beta_2(\tau)$ & $\beta_3(\tau)$ &  $\beta_4(\tau)$ & $\beta_1(\tau)$ & $\beta_2(\tau)$ & $\beta_3(\tau)$ &  $\beta_4(\tau)$ \\ \hline
 \multirow{8}{*}{$\varepsilon$}& \multirow{2}{*}{$N(0,1)$} & $\bar\beta$ &   0.51 & -5.41 & 2.17 & 0.08 & 0.40 & -4.75 & 1.90 & 0.01 & 0.50 & -4.58 & 1.83 & 0.01 & 0.51 & -4.62 & 1.85 & 0.06 & 0.59 & -5.36 & 2.15 & 0.10 
 \\
&&$\sigma(\beta)$ & 0.33 & 0.36 & 0.21 & 0.10 & 0.29 & 0.20 & 0.18 & 0.04 & 0.23 & 0.20 & 0.17 & 0.04 & 0.23 & 0.20 & 0.16 & 0.09 & 0.27 & 0.38 & 0.23 & 0.11 
 \\ 
& \multirow{2}{*}{$t_3$} & $\bar\beta$ & 0.56 & -5.49 & 2.18 & 0.08 & 0.41 & -4.84 & 1.93 & 0.01 & 0.52 & -4.61 & 1.83 & 0.01 & 0.52 & -4.63 & 1.84 & 0.05 & 0.59 & -5.20 & 2.07 & 0.10 
\\
&&$\sigma(\beta)$ & 0.33 & 0.38 & 0.22 & 0.10 & 0.32 & 0.21 & 0.17 & 0.03 & 0.23 &0.21 & 0.16 & 0.03 & 0.23 & 0.21 & 0.16 & 0.08 & 0.27 & 0.37 & 0.20 & 0.10 
 \\
& \multirow{2}{*}{$\mbox{Exp}(2)$} & $\bar\beta$ & 0.42 & -5.48 & 2.19 & 0.08 & 0.46 & -4.84 & 1.93 & 0.00 & 0.53 & -4.69 & 1.87 & 0.00 & 0.53 & -4.77 & 1.91 & 0.06 & 0.59 & -5.32  & 2.13 &  0.12 
\\
&&$\sigma(\beta)$ & 0.35 & 0.38 & 0.21 & 0.10 & 0.28 & 0.21 & 0.19 & 0.03 & 0.23 & 0.22 & 0.17 & 0.03 & 0.23 & 0.20 & 0.16 & 0.08 & 0.26 & 0.34 & 0.20 &0.10 
 \\ 
& \multirow{2}{*}{$\mbox{Cauchy}(0,1)$} & $\bar\beta$ & 0.58 & -5.56 & 2.14 & 0.10 & 0.39 & -4.99 & 1.92 & 0.01 & 0.51 & -4.64 & 1.79 & 0.00 & 0.51 &-4.63 & 1.79 & 0.00 & 0.56 & -5.04 & 1.94 & 0.10 
\\ 
&& $\sigma(\beta)$ & 0.27 & 0.28 & 0.20 & 0.09 & 0.31 & 0.18 & 0.19 & 0.05 & 0.18 & 0.19 & 0.17 & 0.03 & 0.19 & 0.17 & 0.14 & 0.04 & 0.20 & 0.31 & 0.16 & 0.09 
\\ \hline 
  \multicolumn{23}{c}{Model 9}\\ \hline
 & &$\tau$ & \multicolumn{4}{c}{$0.1$} & \multicolumn{4}{c}{$0.3$} &
 \multicolumn{4}{c}{$0.5$} & \multicolumn{4}{c}{$0.7$} & \multicolumn{4}{c}{$0.9$}\\ \hline 
 && & $\beta_1(\tau)$ & $\beta_2(\tau)$ & $\beta_3(\tau)$ &  $\beta_4(\tau)$ &$\beta_1(\tau)$ & $\beta_2(\tau)$ & $\beta_3(\tau)$ &  $\beta_4(\tau)$&$\beta_1(\tau)$ & $\beta_2(\tau)$ & $\beta_3(\tau)$ &  $\beta_4(\tau)$&$\beta_1(\tau)$ & $\beta_2(\tau)$ & $\beta_3(\tau)$ &  $\beta_4(\tau)$ & $\beta_1(\tau)$ & $\beta_2(\tau)$ & $\beta_3(\tau)$ &  $\beta_4(\tau)$ \\ \hline
 \multirow{8}{*}{$\varepsilon$}& \multirow{2}{*}{$N(0,1)$} & $\bar\beta$ &  -0.04 & -4.93 & 0.00 & 0.00 & -0.03 & -4.63 & 0.00 & 0.00 & -0.03 & -4.88 & 0.00 & 0.00 & -0.05 & -5.16 & 0.00 & 0.00 & -0.05 & -5.28 & 0.00 & -0.01 
 \\
&&$\sigma(\beta)$ & 0.21 & 0.25 & 0.08 & 0.07 & 0.16 & 0.21 & 0.06 & 0.06 & 0.15 & 0.70 & 0.06 & 0.06 & 0.17 & 0.31 & 0.07 & 0.07 & 0.20 & 0.50 & 0.08 & 0.07 
\\ 
& \multirow{2}{*}{$t_3$} & $\bar\beta$ & -0.03 & -5.02 & 0.00 & 0.00 & -0.02 & -4.76 & 0.00 & 0.00 & -0.02 & -5.00 & 0.00 & 0.00 & -0.03 & -5.16 & 0.00 & 0.00 & -0.03 &-5.10 & 0.00 & 0.00 
 \\
&&$\sigma(\beta)$ & 0.24 & 0.28 & 0.09 & 0.08 & 0.19 & 0.23 & 0.08 & 0.06 & 0.18 & 0.56 & 0.07 & 0.07 & 0.21 & 0.33 & 0.08 & 0.07 & 0.21 & 0.53 & 0.08 & 0.07
 \\
& \multirow{2}{*}{$\mbox{Exp}(2)$} & $\bar\beta$ & -0.01 & -4.80 & 0.00 & 0.01 &
-0.02 & -4.79 & 0.00 & 0.00 & -0.02 & -4.98 & -0.01 & 0.00 & 0.53 & -4.77 & 1.91 & 0.06 & -0.03 & -5.30 & -0.01 & 0.00 
\\
&&$\sigma(\beta)$ & 0.20 & 0.27 & 0.08 & 0.07 & 0.15 & 0.21 & 0.07 & 0.05 & 0.16 & 0.85 & 0.06 & 0.06 & 0.23 & 0.20 & 0.16 & 0.08 & 0.19 & 0.52 & 0.08 & 0.07 
 \\ 
& \multirow{2}{*}{$\mbox{Cauchy}(0,1)$} & $\bar\beta$ & -0.04 & -5.60 & 0.00 & 0.00 & -0.04 & -5.08 & 0.00 & 0.00 & -0.04 & -5.92 & -0.01 & 0.00 & -0.03 & -5.40 & -0.01 & 0.00 & -0.03 & -4.95 & -0.01 & 0.00 
 \\ 
&& $\sigma(\beta)$ & 0.22 & 0.62 & 0.11 & 0.07 & 0.17 & 0.48 & 0.09 & 0.06 & 0.18 & 0.70 & 0.09 & 0.07 & 0.16 & 0.78 & 0.10 & 0.07 & 0.19 & 1.15 & 0.10  & 0.07 
\\ \hline 
\end{tabular}\label{table: gam fit summary models 7-9}
     \end{adjustbox}
    \egroup
 \end{table}

\subsubsection{High-dimensional Settings}
Since the number of features $p=500$ is relatively large in the high dimensional settings, the available nonparametric estimators are limited. We consider an additive model with polynomial basis up to order $3$, which leads to a 1500 basis expansion terms and 1000 observations situation and require regularization. We choose the minimax concave penalty \cite{zhang2010nearly} in order to obtain relatively consistent estimators of $h(x)$ and demonstrate the performance of Algorithm~\ref{agthm: stepwise backward algorithm}. One typical issue of regularized estimators is that they often produce noisy estimators, i.e., some irrelevant features have near-zero effects. Algorithm~\ref{agthm: stepwise backward algorithm} will largely reduce the number of near-zero effects automatically. We test the proposed method on Models~1-3 in Section~\ref{ssec: low dim settings}. In addition to the empirical mean and standard deviation of the relevant features, we also report in Table~\ref{table: regularization models} the reduction of the ``near-zero" partial effect for the irrelevant features which is defined as 
\begin{equation}
  \mbox{prun}(\tau) = \frac{1}{R (p-2)} (\sum_{j =3}^p I\{\hat \beta_{{\rm init}, j}(\tau) \neq 0 \} - \sum_{j =3}^p I\{\hat \beta_j(\tau) \neq 0 \} ) \times 100\%.
\end{equation}

\begin{table}[!htb]
	\centering
 \caption{The empirical mean and standard deviation of $\hat\beta(\tau)$ for models 1-3 in the high dimensional settings. The estimator $\hat h(x)$ is a regularized additive polynomial fit.}
	\bgroup
	\def\arraystretch{1.2}
	\begin{adjustbox}{max width=\textwidth}
	\begin{tabular}{c |c| c| c c c | c c c  |c c c | c c c|c c c| } \hline
  \multicolumn{18}{c}{Model 1}\\ \hline
 & &$\tau$ & \multicolumn{3}{c}{$0.1$} & \multicolumn{3}{c}{$0.3$} &
 \multicolumn{3}{c}{$0.5$} & \multicolumn{3}{c}{$0.7$} & \multicolumn{3}{c}{$0.9$}\\ \hline 
 && & $\beta_1(\tau)$ & $\beta_2(\tau)$ & $\mbox{prun}(\tau)$  &$\beta_1(\tau)$ & $\beta_2(\tau)$ & $\mbox{prun}(\tau)$ &  $\beta_1(\tau)$ & $\beta_2(\tau)$ & $\mbox{prun}(\tau)$ & $\beta_1(\tau)$ & $\beta_2(\tau)$ & $\mbox{prun}(\tau)$ &  $\beta_1(\tau)$ & $\beta_2(\tau)$ & $\mbox{prun}(\tau)$ \\ \hline
 \multirow{8}{*}{$\varepsilon$}& \multirow{2}{*}{$N(0,1)$} & $\bar\beta$ & 5.80 & -0.03 & 0.72 & 4.20 & -5.31 & 0.68 & 4.41 & -5.27  & 0.76 & 3.55 & -5.01 & 0.70 &  3.24 & -5.12 & 0.68
  \\
  &&$\sigma(\beta)$ & 2.46 & 0.34 & - &  0.45 & 0.19 & - & 0.92 & 0.46 & - & 0.37 & 0.26 & - & 0.28 & 0.20 & - \\ 
& \multirow{2}{*}{$t_3$} & $\bar\beta$ & 5.52 & -0.56  & 0.66 & 4.56 & -5.58 & 0.62 & 4.63 & -5.41 & 0.62 & 3.34 & -4.85 & 0.56 & 3.22 & -5.19 & 0.61
\\
&&$\sigma(\beta)$ &  0.71 & 1.53 & - &  0.19 & 0.14 & - & 0.13 & 0.12 & - & 0.14 & 0.13 & - & 0.09 & 0.14 & - \\
& \multirow{2}{*}{$\mbox{Exp}(2)$} & $\bar\beta$ & 5.52 & -0.56 & 0.66 & 4.56 & -5.58 & 0.62 & 4.63 & -5.41 & 0.62 & 3.34 & -4.85 & 0.56 & 
 3.22 & -5.19 & 0.61
\\
&&$\sigma(\beta)$ & 0.71 & 1.53 & - & 0.19 & 0.14 & - & 0.13 & 0.12 & - & 0.14 & 0.13 & - &  0.09 & 0.14 & - \\ 
& \multirow{2}{*}{$\mbox{Cauchy}(0,1)$} & $\bar\beta$ & 11.53 & -2.11 & 0.96 & 3.45 & -7.24 & 1.02 & 7.41 & -10.03 & 0.91 & 5.49 & -2.92 & 1.03 &  6.75 & -6.85 & 1.00
  \\
&&$\sigma(\beta)$ & 77.73 & 3.08 & - & 15.14 & 6.95 & - &  6.37 & 11.57 & - & 11.37 & 6.78 & - & 12.88 & 10.26 & - \\ \hline 
  \multicolumn{18}{c}{Model 2}\\ \hline
 & &$\tau$ & \multicolumn{3}{c}{$0.1$} & \multicolumn{3}{c}{$0.3$} &
 \multicolumn{3}{c}{$0.5$} & \multicolumn{3}{c}{$0.7$} & \multicolumn{3}{c}{$0.9$}\\ \hline 
&& & $\beta_1(\tau)$ & $\beta_2(\tau)$ & $\mbox{prun}(\tau)$  &$\beta_1(\tau)$ & $\beta_2(\tau)$ & $\mbox{prun}(\tau)$ &  $\beta_1(\tau)$ & $\beta_2(\tau)$ & $\mbox{prun}(\tau)$ & $\beta_1(\tau)$ & $\beta_2(\tau)$ & $\mbox{prun}(\tau)$ & $\beta_1(\tau)$ & $\beta_2(\tau)$ & $\mbox{prun}(\tau)$ \\ \hline
 \multirow{8}{*}{$\varepsilon$}& \multirow{2}{*}{$N(0,1)$} & $\bar\beta$ & 0.48 & -4.69 & 6.56 & 0.15 & -4.86 & 6.38 & 0.03 & -4.91 & 6.47 &  0.01 &  -5.10 & 6.18 & -0.02 & -5.18 & 6.28
 \\
&&$\sigma(\beta)$ & 0.19 & 0.18 & - & 0.26 & 0.15 & - &  0.12 & 0.85 & - & 0.04 & 0.16 & - &  0.08 & 0.24 & -  \\ 
& \multirow{2}{*}{$t_3$} & $\bar\beta$ & 0.53  & -4.64 & 7.04 &  0.20 & -4.87 & 7.31 &  0.00 & -5.04 & 7.24 &  0.00 & -5.12 & 7.18 & -0.02 & -5.05 & 7.23
\\
&&$\sigma(\beta)$ & 0.21 & 0.32 & - &  0.29 & 0.28 & - & 0.04 & 0.35 & - & 0.01 & 0.13 & - & 0.16 & 0.37 & - \\
& \multirow{2}{*}{$\mbox{Exp}(2)$} & $\bar\beta$ & 0.52 & -4.54 & 5.82 & 0.18 & -4.86 & 5.87 &  0.00 & -4.39 &  5.70 &  0.00 & -5.23  & 5.70 &  0.00 & -5.01 & 5.75
 \\
&&$\sigma(\beta)$ & 0.32 & 0.43 & -  & 0.30 &  0.28 & - & 0.09 & 1.74 & -  & 0.03 & 0.15 & - &  0.03 & 0.21 & -
\\ 
& \multirow{2}{*}{$\mbox{Cauchy}(0,1)$} & $\bar\beta$ & 0.47 & -4.51 & 1.25 &  0.37  & -4.80 & 1.30 &  0.00 & -4.81 & 1.20 & -0.01 & -4.95 & 1.19 & -0.02 & -4.62 & 1.26 
\\ 
&& $\sigma(\beta)$ & 0.30 & 0.48 & - & 0.31 & 0.54 & - &  0.00 & 0.49 & - & 0.13 & 0.41 & - & 0.26 & 0.46 & - \\ \hline 
  \multicolumn{18}{c}{Model 3}\\ \hline
 & &$\tau$ & \multicolumn{3}{c}{$0.1$} & \multicolumn{3}{c}{$0.3$} &
 \multicolumn{3}{c}{$0.5$} & \multicolumn{3}{c}{$0.7$} & \multicolumn{3}{c}{$0.9$}\\ \hline 
 && & $\beta_1(\tau)$ & $\beta_2(\tau)$ & $\mbox{prun}(\tau)$  &$\beta_1(\tau)$ & $\beta_2(\tau)$ & $\mbox{prun}(\tau)$ &  $\beta_1(\tau)$ & $\beta_2(\tau)$ & $\mbox{prun}(\tau)$ & $\beta_1(\tau)$ & $\beta_2(\tau)$ & $\mbox{prun}(\tau)$ &  
 $\beta_1(\tau)$ & $\beta_2(\tau)$ & $\mbox{prun}(\tau)$ \\ \hline
 \multirow{8}{*}{$\varepsilon$}& \multirow{2}{*}{$N(0,1)$} & $\bar\beta$ & 0.69 & -4.93 & 0.53 & 0.34 & -5.11 & 0.61 & 0.09 & -5.12 & 0.68 & -0.33 & -5.05 & 0.63 & -0.96 & -4.86 & 0.66
  \\
&&$\sigma(\beta)$ & 0.11 & 0.20 & - & 0.11 & 0.14 & - & 0.05 & 0.16 & - &  0.10 & 0.20 & - & 0.12 & 0.20 & -\\ 
& \multirow{2}{*}{$t_3$} & $\bar\beta$ & 0.74 & -5.02 & 0.59 & 0.36 & -5.11 & 0.47 &  0.13 & -5.16 & 0.58 & -0.33 & -5.00 & 0.52 & -0.99 &-4.83 & 0.60 
 \\
&&$\sigma(\beta)$ & 0.08 & 0.18 & - & 0.10 & 0.13 & - & 0.06 & 0.12 & - &  0.07 & 0.13 & - & 0.06 & 0.21 & -  \\
& \multirow{2}{*}{$\mbox{Exp}(2)$} & $\bar\beta$ & 0.64 & -4.92 & 0.61 &  0.35 & -5.12 & 0.53 & 0.10 & -5.10 & 0.57 & -0.30 & -4.89 & 0.56 & -0.97 & -5.07 & 0.52   
\\
&&$\sigma(\beta)$ & 0.16 & 0.20 & - & 0.09 & 0.15 & - & 0.05 & 0.14 & - & 0.07 & 0.14 & - &  0.07 & 0.20 & -  \\ 
& \multirow{2}{*}{$\mbox{Cauchy}(0,1)$} & $\bar\beta$ & 0.80 & -6.01  & 0.20 & 0.69 & -5.30  & 0.05 &  0.33 & -5.44 & 0.18 & -0.34 & -6.82 & 0.13 & -1.42 & -5.02 & 0.11 
 \\ 
&& $\sigma(\beta)$ & 0.07 & 0.49 & - & 0.09 & 0.16 & - &  0.06 &  0.28 & - & 0.06 & 0.62 & - & 0.16 & 0.33 & -
 \\ \hline 
\end{tabular}
\label{table: regularization models}
 \end{adjustbox}
    \egroup
 \end{table}

From Tables~\ref{table: gam fit summary models 1-3}, \ref{table: gam fit summary model 4-6}, and \ref{table: regularization models}, the proposed method does not have stable performance under $\mbox{Cauchy}(0,1)$ distributed error. This is under the expectation since the moments of $\mbox{Cauchy}(0,1)$ errors do not exist; this can cause issues when optimizing the loss functions. In the low dimensional settings, the stepwise backward pruning in Algorithm~\ref{agthm: stepwise backward algorithm} has stable performance which is reflected by both the mean and standard deviation of irrelevant features being zero or near-zero in Tables~\ref{table: gam fit summary models 1-3}, \ref{table: gam fit summary model 4-6}. In the high dimensional settings, the stepwise backward pruning helps reduce the noisy components, which is especially drastic for Model 2.








\subsubsection*{Acknowledgments}
The research presented in this paper was carried out on the High Performance Computing Cluster supported by the Research and Specialist Computing Support service at the University of East Anglia. 
C.~Li is partially supported by a start-up grant from Iowa State University.

\appendix
\section{Appendix}
\label{sec:appendix}

\subsection{Preliminaries and Derivations}

\begin{definition}[Influence function]\label{def:influence-function}
The small change on a distributional statistic $\nu$ by assigning a point mass at $y$ with weight $t$ for $F_Y$ is measured by the influence function \cite{hampel1974influence} calculated as 
\[
\mbox{IF}(y, \nu) = \lim_{t \to 0} \frac{\nu(G_Y) - \nu(F_Y)}{t },
\]
where $G_Y = (1- t) F_Y + t \delta_y$ and $\delta_y  = I\{Y = y\}$.
\end{definition}

\begin{lemma}[von Mises linear approximation]\label{lemma:von-mises}
    Let $G = (1 -t)F + t F'$ where $F$ and $F'$ are two distribution functions and $t \in [0, 1]$. The functional $\nu(G)$ around $F$ can be approximated as 
  \begin{equation}\label{eq: von mises general}
    \nu(G) = \nu(F) + t \int \mbox{IF}(x;\nu) d (G - F)(y) + o_\nu(t),  
  \end{equation}
where the remainder term $o_\nu$ is of order $t$ and depends on the functional $\nu$, and $\mbox{IF}(x; \nu)$ refers to the influence function at $F$.
\end{lemma}


\end{document}